\relax
\makeatletter
\def\@copyrightspace{\relax}
\makeatother
\documentclass[letterpaper]{article}

\usepackage{aaai21}  % DO NOT CHANGE THIS
\usepackage{times}  % DO NOT CHANGE THIS
\usepackage{helvet} % DO NOT CHANGE THIS
\usepackage{courier}  % DO NOT CHANGE THIS
\usepackage[hyphens]{url}  % DO NOT CHANGE THIS
\usepackage{graphicx} % DO NOT CHANGE THIS
\usepackage{graphicx}  % DO NOT CHANGE THIS
\usepackage{natbib}  % DO NOT CHANGE THIS AND DO NOT ADD ANY OPTIONS TO IT
\usepackage{caption} % DO NOT CHANGE THIS AND DO NOT ADD ANY OPTIONS TO IT
\usepackage{url}            % simple URL typesetting
\usepackage{booktabs}       % professional-quality tables
\usepackage{amsfonts}       % blackboard math symbols
\usepackage{amssymb}
\usepackage{bm}
\usepackage{nicefrac}       % compact symbols for 1/2, etc.
\usepackage{microtype}      % microtypozgraphy
\usepackage{amsmath}
\usepackage{xcolor}
\usepackage{graphicx} 
\usepackage{adjustbox}
\usepackage{enumitem}

\usepackage[switch]{lineno}
\usepackage{appendix}

\usepackage[ruled,vlined]{algorithm2e}
\usepackage{dblfloatfix}
\usepackage{afterpage}
\usepackage{tabularx}
\usepackage{multirow}
\usepackage{hhline}
\usepackage{hyperref}

\newcolumntype{Y}{>{\centering\arraybackslash}X}

\newcommand{\tacredtrain}{\texttt{TACRED\textsubscript{train}}}
\newcommand{\retacredtrain}{\texttt{Re-TACRED\textsubscript{train}}}
\newcommand{\tacredtest}{\texttt{TACRED\textsubscript{test}}}
\newcommand{\retacredtest}{\texttt{Re-TACRED\textsubscript{test}}}

\title{Re-TACRED: Addressing Shortcomings of the TACRED Dataset}

\author {
        George Stoica \textsuperscript{\rm \thanks{Work performed while at Carnegie Mellon University.}},
        Emmanouil Antonios Platanios \textsuperscript{\rm 2},
        Barnabas Poczos \textsuperscript{\rm 1} \\
}
\affiliations {
    \textsuperscript{\rm 1} Carnegie Mellon University \\
    \textsuperscript{\rm 2} Microsoft Semantic Machines \\
    gstoica27@gmail.com, emplata@microsoft.com, bapoczos@cs.cmu.edu
}

\begin{document}

\maketitle
% \linenumbers

\begin{abstract}

TACRED is one of the largest and most widely used sentence-level relation extraction datasets.
Proposed models that are evaluated using this dataset consistently set new state-of-the-art performance.
However, they still exhibit large error rates despite leveraging external knowledge and unsupervised pretraining on large text corpora.
A recent study suggested that this may be due to poor dataset quality.
The study observed that over $50\%$ of the most challenging sentences from the development and test sets are incorrectly labeled and account for an average drop of $8\%$ f1-score in model performance.
However, this study was limited to a small biased sample of 5k (out of a total of 106k) sentences, substantially restricting the generalizability and broader implications of its findings.
In this paper, we address these shortcomings by: (i) performing a comprehensive study over the whole TACRED dataset, (ii) proposing an improved crowdsourcing strategy and deploying it to re-annotate the whole dataset, and (iii) performing a thorough analysis to understand how correcting the TACRED annotations affects previously published results.
After verification, we observed that $23.9\%$ of TACRED labels are incorrect.
Moreover, evaluating several models on our revised dataset yields an average f1-score improvement of $14.3\%$ and helps uncover significant relationships between the different models (rather than simply offsetting or scaling their scores by a constant factor).
% Perhaps one of our most striking findings is that 
% about $22.9\%$ of the TACRED labels are incorrect, with an additional $15.4\%$ involving incorrect type annotations (e.g., confusing \texttt{ people} with \texttt{ organizations}). Moreover, evaluating several models on our revised dataset yields an average F1 improvement of $12.3\%$.
% with $15.4\%$ of them involving incorrect type annotations (e.g., confusing \texttt{  people} with \texttt{organizations}).
Finally, aside from our analysis we also release \textbf{Re-TACRED}, a new completely re-annotated version of the TACRED dataset that can be used to perform reliable evaluation of relation extraction models.

% \citet{tacredReannotate} previously suggested that this may be due to poor dataset quality.
% They observed that over $50\%$ of the most challenging sentences from the development and test sets are incorrectly labeled and account for an average drop of $8\%$ F1-score in model performance.
% However, their study was limited to a small biased sample of 5k (out of a total of 106k) sentences, substantially restricting the generalizability and broader implications of their findings.
% In this paper, we address these shortcomings by: (i) performing a comprehensive study over the whole TACRED dataset, (ii) proposing an improved crowdsourcing strategy and deploying it to re-annotate the whole dataset, and (iii) performing a thorough analysis to understand how correcting TACRED affects previously published results.
% After verification, we observe that $23.9\%$ of TACRED labels are incorrect.
% Moreover, evaluating several models on our revised dataset yields an average f1-score improvement of $14.3\%$ and also helps uncover stronger relationships between the different models (rather than simply offsetting or scaling their scores by a constant factor).
% Finally, aside from our analysis we also release \textbf{Re-TACRED}, a new and completely re-annotated version of the TACRED dataset that can be used to perform reliable evaluation of relation extraction models.

\end{abstract}

\section{Introduction}
\label{sec:introduction}

Many applications ranging from medical diagnostics to search engines rely on the ability to uncover relationships between diverse concepts.
% seemingly disparate concepts based on existing knowledge.
Relation extraction (RE) is a popular learning task aimed at extracting such relationships between entities in plain text.
For example, given the sentence ``\texttt{[William Shakespeare]\textsubscript{SUB} was born in [England]\textsubscript{OBJ}},'' where ``\texttt{William Shakespeare}'' and ``\texttt{England}'' are the subject and object entities respectively, the objective of a RE method is to infer the correct relation (e.g., \texttt{PERSON:BORN\_IN\_COUNTRY}) between the subject and the object.
Developing successful RE methods requires robust ways to evaluate their qualities. 
TACRED \citep{palstm} is one of the largest and most widely used such datasets. 
% Two important benchmarks are the SemEval 2010 Task 8 \citep{semeval} dataset, and the significantly larger and more widely used TACRED \citep{palstm} dataset.
% However, \citet{tacredReannotate} show that TACRED suffers from flaws that may impact its use as an evaluation benchmark.
% In this paper, we propose a new crowdsourcing task design to re-annotate TACRED, resulting in Re-TACRED, a fully re-annotated version of TACRED.
% The resulting dataset is of significantly higher quality and results in an interesting adjustment of prior results that were obtained using TACRED (Section \ref{sec:analysis} describes this in more detail).

TACRED consists of 106,264 sentences of varied complexity that were annotated using Amazon Mechanical Turk (AMT).
Although just three years-old, a multitude of approaches have been proposed and evaluated using the TACRED dataset.
These approaches typically leverage an assortment of different knowledge: (i) auxiliary named entity recognition (NER) and part-of-speech (POS) tag information \citep[e.g.,][]{palstm,cgcn,aggcn}, (ii) sentence dependency parses \citep[e.g.,][]{cgcn, aggcn}, (iii) fine-tuned pre-trained language representations \citep[e.g.,][]{bert-em, knowbert, tre, spanbert,dgspanbert, ernie}, or (iv) even external training data \citep[e.g.,][]{bert-em, knowbert}.
% , in order to push the state-of-the-art (SoTA) performance on TACRED.
Recently, at the time of writing, methods have converged to $\sim$71.5\% f1-score on the test data, which raises the question of whether we have reached the maximum possible attainable performance on the TACRED dataset, and if so, why?
\citet{tacredReannotate} investigated these questions by performing a comprehensive review of the 5,000 most misclassified TACRED development and test split sentences among 49 existing RE methods.
% , and re-annotation using an improved crowdsourcing task design.
They observed that over 50\% of the sentences were in fact labeled incorrectly, leading to an average model performance improvement of 8\% after correcting these labels.
Furthermore, they identified several error categories that describe model mistakes on their revised test split.
% , and validate them on their revised TACRED test split.
% Overall, they observe that many model errors are attributed to sentences where the subjects are people and the objects are locations, and those where the subjects and objects are the same type of concept (e.g. people).
% In addition, even after revision, the TACRED test split contained $\sim$10,000 uncorrected and noisily labeled sentences.
% Furthermore, they identified several error categories that describe model mistakes on their revised labels. % , and validated them on the full TACRED test split.
% Overall, they observed that the models many model errors are attributed to sentences where the subjects are people and the objects are locations. % \texttt{PERSON:LOCATION} mentions (subjects are people and objects are locations), and those whose entities are equivalent types with existing relations.
However, 
% while \citet{tacredReannotate} exposed several important shortcomings of the TACRED dataset and the types of errors models exhibit over it, 
the broader impact of their work is limited by two key factors.
First, they restricted their dataset revisions to a small and biased sample of TACRED. 
Thus, it is not clear whether their findings would be true for the full TACRED dataset.
% While their corrections yielded significant performance improvements, 
% Perhaps most importantly, this small sample was also biased because they only considered the 5,000 most misclassified TACRED sentences among 49 existing RE methods, which means that these sentences could likely also have been more ambiguous or otherwise difficult than the rest of the TACRED sentences.
Second, even after their revisions, the majority of TACRED remained uncorrected, making it challenging to identify if new errors made by the methods were primarily due to model capacity, data error, or a mixture of both.

% \cite{tacredReannotate} investigate these questions by performing a comprehensive review of the five thousand most misclassified TACRED sentences among $49$ existing RE methods. Leveraging crowd-annotations, they observe that over $50\%$ of sentences are incorrectly labeled, leading to an average model improvement of over $8\%$ after correcting the labels. Furthermore, they identify several error categories that describe model missclassifications on their revised instances, and validate them on the full TACRED test split. Overall, they observe that model many model errors are attributed to sentences with \texttt{PERSON:LOCATION} mentions (subjects are people and objects are locations), and those whose entities are equivalent types with existing relations. 

%  However, while \citet{tacredReannotate} exposed several important shortcomings of the TACRED dataset and the types of errors models exhibit, the broader impact of their work is limited by two factors. First, they restricted their dataset revision to a small and biased sentence sample. While their corrections yielded significant performance improvements, it is difficult the results to expected improvement across the full dataset. Second, their core analysis centered around predominately uncorrected TACRED data. For instance, all methods used were trained on the original noisy TACRED train split, and evaluated on a combination of original and revised labels. This makes it challenging to identify if errors made by methods are primarily due to model capacity, data error, or a mix of them.

In this paper, we aim to address these shortcomings by performing a re-annotation of the entire TACRED dataset.
Our contributions can be summarized as follows:
\begin{itemize}[noitemsep,label=--]
    \item \underline{Annotation:}
        We propose an improved and cost-efficient crowdsourcing annotation strategy that we subsequently deploy to re-annotate the full TACRED dataset.
        Our task design tackles an important flaw in the original TACRED data collection process, refines existing relation definitions to be better suited for the TACRED dataset, and uses quality assurance mechanisms in order to ensure increased annotation quality (and thus accuracy).
        Our annotators achieve an average agreement rate of 82.3\% and an inter-annotator Fleiss' kappa of .77, which is significantly higher than the .54 kappa achieved by \citet{palstm}.
    \item \underline{Analysis:}
        We perform a thorough comparison of the TACRED labels and our new re-annotated labels. We analyze both their qualitative differences, and their impact on the evaluation and comparison of existing RE models.
        Our results show that our corrections significantly improve model performance by an average of 14.3\% f1-score, and also indicate that prior analysis on the types of errors that the models make may have been misguided due to the wrong TACRED labels.
    \item \underline{Dataset Release:}
        We release our newly corrected TACRED labels publicly online (\url{https://github.com/gstoica27/Re-TACRED}). Due to licensing restrictions, we cannot release complete dataset, but similar to \citet{tacredReannotate}, we release a patch that contains all of our revisions.
        We term the corrected dataset Revised-TACRED (Re-TACRED).
\end{itemize}

\section{Background}
\label{sec:background}

The TAC relation extraction dataset (TACRED), introduced by \citet{palstm}, is one of the largest and most widely used datasets for sentence-level relation extraction.
It consists of over 106,000 sentences collected from the 2009-2014 TAC knowledge base population (KBP) evaluations, with those between 2009-2012 used for training, 2013 for development, and 2014 for testing.
Each TACRED instance consists of a sentence and two non-overlapping contiguous spans of text that represent a subject and an object, respectively, each with pre-specified ``types'' (e.g., \texttt{PERSON} or \texttt{CITY}).
Furthermore, each instance is assigned one of 42 labels that describes the relationship between the subject and the object.
These labels consist of 41 relation types that describe the existence of some relationship between the subject and the object (e.g., \texttt{PERSON:CITY\_OF\_BIRTH}), and a special \texttt{NO\_RELATION} predicate to indicate the absence of a relationship.
For example, consider the sentence ``\texttt{[John Doe]\textsubscript{SUB} lives in [Miami]\textsubscript{OBJ}}.''
In this case, the subject is a \texttt{PERSON} and the object is a \texttt{CITY}.
In TACRED, all relations are {\em typed}, meaning that they only apply to a specific subject and object type.
The subject type is always either \texttt{PERSON} or \texttt{ORGANIZATION}, and there exist 17 unique object types.
There are a total of 27 subject-object type pairs with corresponding candidate relations.

Instances in the original TACRED dataset were annotated with labels using the Amazon Mechanical Turk (AMT) crowdsourcing platform.
The AMT workers were provided sentences with their subject and object spans highlighted, and were asked to choose the appropriate label from a set of suggestions (i.e., the annotation task was framed as a multiple choice task).
The suggestions included all labels that were compatible with the subject and object types, along with the special \texttt{NO\_RELATION} label.

\subsection{TACRED Quality}

\citet{palstm} manually verified TACRED annotation quality over a random sample of 300 instances.
They reported that they observed a high annotation accuracy of 93.3\%, with respect to what they considered to be the correct labels for these instances.
Coupled with a moderate Fleiss' kappa of .54 over 761 randomly selected annotation pairs, they assumed an acceptable level of label quality.
However, recent work suggests that the true annotation quality may be significantly lower than previously estimated.
\citet{tacredReannotate} used crowdsourcing to manually verify labels for the five thousand most miss-classified sentences from $49$ existing relation extraction methods.
Their annotation task was designed similar to that of \citet{palstm}, with two primary differences to help identify potential issues. First, only workers with prior training in general linguistics were allowed to participate, and these workers were further pruned by asking them to correctly label 500 manually chosen and hand-labeled sentences from the original TACRED development set.
Second, the set of possible choices presented to the workers also included the set of predictions made by pre-trained (on the original TACRED dataset) relation extraction models. These predictions may have included type-incompatible relations to help identify cases of wrongly-assigned types.
% I skipped the "allowing a sentence to express multiple relations part" because they don't really explain how they do it.
% Following a similar procedure to \citet{palstm}, crowd workers were presented with example sentences with highlighted entity mentions.
% Workers were then asked to assign a label from a set consisting of all type-compatible relations, the original TACRED annotation, and labels predicted by the 49 methods.
% A second label was allowed only if the sentence expressed multiple relationships.
% While rare, sentence syntax can imply multiple relations, e.g. "native" implies both "\texttt{per:born\_in}" and "\texttt{per:lives\_in}."
% In addition, all crowd-workers were previously trained in general linguistics and had passed a trial of 500 selected sentences from the TACRED development set.
Using this re-annotation procedure, they observed that {\em over 50\% of the TACRED annotations in their sample were incorrect}.
Among the wrongly-annotated instances, they found that $36\%$ were erroneously labeled as \texttt{NO\_RELATION}, $49\%$ were incorrectly assigned relations other than \texttt{NO\_RELATION}, and $15\%$ were assigned the wrong label among non-\texttt{NO\_RELATION} labels.
Notably, their revised dataset resulted in an average f1-score improvement of $8\%$ over the unaltered TACRED dataset, suggesting that using TACRED for evaluating methods may potentially result in inaccurate conclusions.
Moreover, their Fleiss' kappa for the new annotations was $0.80$ for the development set and $0.87$ for the test set, suggesting high annotation quality. 
% Note that, these numbers are both significantly higher than the one computed for the original TACRED labels ($0.54$).

% However, recent work by \citet{tacredReannotate} suggests that the annotation quality may be significantly lower than previously estimated. 
% However, recent work by \cite{tacredReannotate} suggests that the annotation quality may be significantly lower than previously estimated. Specifically, \citet{tacredReannotate} performed a comprehensive crowd-sourced verification of the five thousand most missclassified sentences from the development and test sets by 49 different relation extraction models. Similar to \citet{palstm}, crowd workers were given example sentences with highlighted entity mentions, and asked to assign a label amongst a set of candidate relations. These relations consisted of the set of all compatible with sentence entity types, the existing sentence annotation, and the set of inferred relations by the 49 methods. Workers assigned to these tasks were all previously trained in general linguistics and had passed a trial exam of 500 relevant example sentences. The resultant development and test split inter-annotator kappa's were very high at $\kappa_{\text{dev}} = .80$ and $\kappa_{\text{test}} = .87$ respectively. Most notably however, \cite{tacredReannotate} observed that over $50\%$ of the five thousand sentences examined required relabeling, and found that correcting these labels resulted in an over $8\%$ average F1 improvement across relation extraction methods on the TACRED dataset.

While \citet{tacredReannotate} demonstrated several shortcomings of the TACRED dataset, the broader impact of their work is restricted by both their small and biased sample set, and the fact that their analysis was performed over a predominately uncorrected version of the TACRED dataset.
% by two key factors.
% First, their analysis was only performed over a small, non-random, and biased sample of TACRED sentences.
Although correcting this small set of labels yielded significant impact on the evaluation of existing relation extraction models, it is difficult to generalize the results to the full dataset.
% Second, all 49 methods utilized to generate their sample were trained on the original noisy TACRED annotations, meaning that the methods were likely biased (and thus were the selected samples too).
% Errors in development and test partitions indicate likely issues with train labels as well, biasing models towards certain types of predictions and misclassifications.
% Second, all $49$ methods were trained on the original noisy TACRED annotations.
% Thus, it is not clear whether the remaining model misclassifications are due to further underlying data inaccuracies, or model capability, and a few questions remain that need to be answered.
These disadvantages raise several questions that are difficult to answer with their study.
Can we design a cost-effective yet robust crowdsourcing annotation task in order to correct all of the TACRED dataset and allow the research community to benefit from more accurate evaluations of new methods?
Can we expect similar performance improvements when re-annotating the full dataset?
How do model errors change when using a fully re-annotated dataset?
These questions form the main motivation for this paper.
% are difficult to answer based on the work of \citet{tacredReannotate} and 
% form our main motivation for the work described in this paper.

% Do we expect to see similar error rates and or improvements after verifying the full dataset? Are the types of inaccuracies present in the data mirrored or similar to what is observed in the five thousand most challenging sentences? These questions are difficult to answer with a non-representative sample.
% While \cite{tacredReannotate} illustrate and empirically demonstrate the potential shortcomings of TACRED labels, the broader impact of their analysis is limited by two factors.
% First, their verification sample is less than $5\%$ of the dataset size, and is biased as it arguably contains only the most ambiguous and difficult TACRED sentences. Thus, their resultant label error rate is likely substantially larger than the true TACRED label error rate and cannot serve as a representation of TACRED's quality as a whole. 
% Second, they evaluate model performance differences based on an unaltered and uncorrected TACRED training split. This makes it difficult to generalize their results to the expected performance improvements from correcting the full dataset.

\section{TACRED Revision}
We propose a new crowdsourcing task design that improves upon previous approaches along the following directions:
\begin{enumerate}[noitemsep]
  \item \underline{Wrong Type Handling:}
    We performed a manual analysis of 1,000 randomly selected instances and found that about 5\% of them have incorrect types for the subject, the object, or both (e.g., ``Thomas More Law Center'' tagged as a \texttt{PERSON} instead of an \texttt{ORGANIZATION}).
    % This is important because the task design of both \citet{palstm} and \citet{tacredReannotate} only presented the annotators candidate relations that matched the pre-specified subject and object types.
    This is important because the task design of \citet{palstm} only presented the annotators candidate relations that matched the pre-specified subject and object types.
    Therefore, if the types were wrong, the annotators had no possible way of choosing the right relation.
    % and often resorted to the \texttt{NO\_RELATION} answer.
    % This resulted in a dataset that is heavily biased towards \texttt{NO\_RELATION} (as the authors also point out).
    In Section~\ref{sec:wrong-type-handling}, we propose a cost-effective modification to the previous task design that addresses this issue.
    % a modification to the previous task design that addresses this issue in a cost-effective manner.
  \item \underline{Relation Definition Refinements:}
    Similar to \citet{palstm} and \citet{tacredReannotate}, we initially defined all possible relations according to the TAC KBP documentation (available at \url{https://tac.nist.gov/2017/KBP/index.html}).
    However, we observed that the documentation is ambiguous or unintuitive in a small number of cases.
    This leads to worker confusion and poor annotation quality.
    % This results in confusing the annotators and thus in bad annotation quality.
    We address this problem by altering problematic relation definitions, described in Section~\ref{sec:relation-definitions-refinement}.
  \item \underline{Quality Assurance:}
    In order to ensure high-quality annotations, we employ a two-step quality assurance process for our annotators. This is described in Section~\ref{sec:quality-assurance}.
  \item \underline{Miscellaneous Revisions:}
    During our quality analysis for TACRED, we discovered sentences that were not written in the English language.
    Moreover, we analyze the sentences which posed the greatest challenges for our workers. 
    We address these issues in Section~\ref{sec:miscellaneous-revisions}. 
\end{enumerate}
The following sections describe our overall crowdsourcing task design, as well as our approach along each of these directions in detail.
Note that we {\em re-annotate the full TACRED dataset} using the Amazon Mechanical Turk (AMT) platform. (as opposed to a small fraction of it like \citet{tacredReannotate}).
Finally, in Section~\ref{sec:analysis} we perform an analysis of the resulting changes in the TACRED dataset and their impact in evaluating existing relation extraction methods.

\subsection{Task Design}
\label{sec:task-design}

Labeling TACRED is challenging due to its large size and complex structure.
Sentences contain variable amounts of syntactic and lexical ambiguity, making it difficult for crowd-workers to identify the right relation among 42 choices.
% Furthermore, the intended meaning of certain relations is unclear.
In order to reduce annotation complexity, we follow a similar approach to \citet{palstm, tacredReannotate}.
We first group TACRED sentences based on their corresponding subject and object types (e.g., the sentence ``\texttt{[Holly]\textsubscript{SUB} showed off [her]\textsubscript{OBJ} jewelry}'' is grouped together with sentences whose subject and object both have type \texttt{PERSON}), and we then assign each group a filtered candidate set of labels that consists only of relations that are {\em type-compatible} (e.g., relations between people), along with the special \texttt{NO\_RELATION} label.
Given that the provided types may be wrong (as mentioned earlier), we also allow the annotators to select a special \texttt{WRONG\_TYPES} label for each instance.
This is because, if either of the types is incorrect, then the candidate label set may no longer be truly type-compatible, thus only providing implausible options to the annotators.
This ought to further reduce confusion in cases when the types are incorrect, because annotators are made explicitly aware of this possibility and are provided with an option for them.
% In the following three sections we provide details on three additional important aspects of our task design.

\subsection{Wrong Type Handling}
\label{sec:wrong-type-handling}
The inclusion of the \texttt{WRONG\_TYPE} label implies that the original candidate label sets of affected sentences are not compatible with the sentence. 
To find the correct relation, each sentence must be re-labeled according to different label sets until a match is found. 
% The inclusion of the new \texttt{WRONG\_TYPE} relation means that now, whenever our annotators choose that as the answer, we need to re-annotate the corresponding sentence using a larger set of possible candidate relations that are not compatible with original (wrong) types.
% A potential approach to this problem would be to consider all possible pairs of subject and object types and start going through them until the annotators agree on a relation other than \texttt{WRONG\_TYPE}.
A potential approach to this problem would be to iteratively consider all possible pairs of subject and object types until annotators agree on a relation other than \texttt{WRONG\_TYPE}.
However, such a solution would be prohibitively expensive as in the worst case 27 separate annotation tasks would need to be performed for a single sentence.
If just 5\% of TACRED sentences have wrong types (our estimate based on a $1,000$ sentence sample), then the worst case annotation cost would increase by $\sim$130\%.
% However, in the worst case this would result in 27 separate annotation tasks for a single sentence, which is prohibitive in terms of cost.
% Specifically, if just 5\% of the TACRED sentences have wrong types (which is our estimate based on a 1,000 sentence sample), then the overall annotation cost would increase by $\sim$130\%.

\begin{table*}[!b]
    \centering
    \scriptsize
	\setlength{\tabcolsep}{3pt}
    \renewcommand{\arraystretch}{0.8}
	\begin{tabularx}{\textwidth}{|r|c|X|}
		\hhline{|-|-|-|}
		\textbf{Super-Cluster} & \textbf{Subject Type} & \textbf{Object Types} \\
		\hhline{:=:=:=:}
		\texttt{org2miscmulti} & \multirow{4}{*}{\texttt{ORGANIZATION}} & \texttt{URL}, \texttt{DATE}, \texttt{NUMBER}, \texttt{RELIGION}, \texttt{IDEOLOGY}, \texttt{MISC} \\
		\texttt{org2locmulti} & & \texttt{CITY}, \texttt{COUNTRY}, \texttt{STATE\_OR\_PROVINCE}, \texttt{LOCATION} \\
		\texttt{org2org} & & \texttt{ORGANIZATION} \\
		\texttt{org2per} & & \texttt{PERSON} \\
		\hhline{|-|-|-|}
		\texttt{per2miscmulti} & \multirow{4}{*}{\texttt{PERSON}} & \texttt{TITLE}, \texttt{DATE}, \texttt{CRIMINAL\_CHARGE}, \texttt{RELIGION}, \texttt{NUMBER}, \texttt{CAUSE\_OF\_DEATH}, \texttt{DURATION}, \texttt{MISC} \\
		\texttt{per2locmulti} & & \texttt{NATIONALITY}, \texttt{COUNTRY}, \texttt{STATE\_OR\_PROVINCE}, \texttt{CITY}, \texttt{LOCATION} \\
		\texttt{per2org} &  & \texttt{ORGANIZATION} \\
		\texttt{per2per} & & \texttt{PERSON} \\
		\hhline{|-|-|-|}
	\end{tabularx}
    \caption{
        Mappings between super-clusters and sentence groups. 
        Sentence groups are defined by the pair, (\texttt{SUBJECT\_TYPE}, \texttt{OBJECT\_TYPE}), which describes the subject and object type of all sentences in the group.
        The leftmost column denotes each super-cluster name. 
        The middle column lists the two possible subject types (\texttt{ORGANIZATION} and \texttt{PERSON}), while the rightmost column shows the list of object types whose pairing with the corresponding subject type is an element of the respective super-cluster. For instance, (\texttt{PERSON}, \texttt{TITLE}) represents the sentence group where all sentence subject types are \texttt{PERSON} and all object types are \texttt{TITLE}. From the table, this group is an element of the \texttt{per2miscmulti} super-cluster.
    }
    \label{tab:cluster_mappings}
\end{table*}

We address this issue by defining 8 {\em super-clusters} over relations, such that each super-cluster contains at least one sentence group (i.e., sentences that correspond to a specific subject-object type pair),
% , that does not overlap with any other super-cluster.
and every sentence group belongs to exactly one super-cluster.
To illustrate, let one such super-cluster describe all sentences that exhibit a relationship between a \texttt{PERSON} and a \texttt{LOCATION}.
Sentence groups within this cluster describe different aspects of the super-cluster relationship (e.g., relationships between people and cities), and do not appear in any other super-cluster.
% Moreover, it is guaranteed that each sentence group in this super-cluster does not appear in any others.
We specify each cluster by aggregating sentence groups whose types were most confused with one another from a random sample of a 1,000 sentences.
% our preliminary investigation over a 1,000-sized random sample of TACRED sentences.
Our final super-clusters are shown in Table \ref{tab:cluster_mappings}.% \ref{app:super-clusters}
We define each cluster's candidate label set as the union of the candidate sets for each of its sentence group members.
This increases the probability that type-compatible relations exist for incorrectly-typed sentences within a super-cluster.
Moreover, this approach reduces the worst-case overall annotation cost by a factor of $27 / 8 \approx 3.4$.
% However, we hypothesize that the cost reduction will be higher because the super-clusters are constructed based on the most often confused types and are thus likely to contain the correct relation for cases where the subject and object types are incorrect.

However, our modified ``super-cluster''-based sentence aggregation also increases the size of the candidate label set presented to workers during annotation.
While in many cases the resultant set is reasonably sized (under 9 relations), a minority of clusters have very large label sets, containing up to 14 relations.
Large label sets can make it challenging for annotators to accurately and efficiently choose the most appropriate answer.
To ensure that the candidate sets we present to annotators are not too large, we impose a maximum size of 9 relations for each sentence.
Clusters with corresponding label sets of size less than or equal to 9 are left intact and are annotated in a {\em single-stage} fashion.
Larger clusters, however, are broken down into sub-clusters and are annotated using a {\em multi-stage} process.
The single-stage annotation process consists of asking a single question for each sentence, where the candidate set of relations contains all of the corresponding super-cluster relations.
The multi-stage annotation process consists of splitting a large cluster's label set into subsets such that each subset has fewer relations than our threshold (i.e., 9).
Then, one of these subsets is selected and annotated in the same way as for the single-stage process.
Afterwards, all sentences assigned to the special \texttt{WRONG\_TYPE} relation (indicating that none of the relations in the candidate subset were plausible) are re-annotated using a different subset of relations.
This process is repeated until either all of the subsets are exhausted, or all of the sentences are annotated with labels other than the special \texttt{WRONG\_TYPE} relation.

\subsection{Relation Definition Refinements}
\label{sec:relation-definitions-refinement}

In this section, we describe the changes we made to the original TAC KBP relation definitions in order to make them more clear and intuitive.
% and remove any potential sources of confusion for the annotators.

\vspace{1ex}\noindent\underline{\texttt{PERSON:IDENTITY}:}
We observed substantial inconsistencies in TACRED between the relations \texttt{PERSON:OTHER\_FAMILY} and \texttt{NO\_RELATION} in sentences whose subject and object refer to the same person in a pronominal manner (e.g., ``\texttt{[Holly]\textsubscript{SUB} shows off a few pieces of [her]\textsubscript{OBJ} jewelry line here},'' where the subject and object are denoted as described in Section \ref{sec:introduction}).
% Note, \texttt{per:alternate\_names} is not an appropriate label because it only allows references to the same person using different names.
Despite accounting for nearly 10\% of TACRED, these sentences are difficult to annotate due to ambiguity in the TAC KBP label guidelines.
% To this end, we decided to merge the \texttt{PERSON:OTHER\_FAMILY} and \texttt{PERSON:ALTERNATE\_NAMES} relations.
To this end, we extended the definition of a similar relation \texttt{PERSON:ALTERNATE\_NAMES} to include any pronominal identity references. 
% To this end, we opted to include these types of relationships in the \texttt{PERSON:ALTERNATE\_NAMES} relation.
% Namely, we extended the definition of \texttt{PERSON:ALTERNATE\_NAMES} to also explicitly account for references to the same person, instead of only references using {\em different names}.
Furthermore, in order to avoid confusion and incompatibilities between TACRED and Re-TACRED (our improved TACRED dataset), we renamed the \texttt{PERSON:ALTERNATE\_NAMES} to \texttt{PERSON:IDENTITY}. 
Additional details can be found in Appendix \ref{app:relation-alterations}.

\vspace{1ex}\noindent\underline{\texttt{ORGANIZATION:\{MEMBER\_OF/MEMBERS\}}:}
% The relations \texttt{ORGANIZATION:MEMBER\_OF} and \texttt{ORGANIZATION:PARENTS}, and their corresponding inverses, \texttt{ORGANIZATION:SUBSIDIARIES} and \texttt{ORGANIZATION:MEMBERS}, describe the relationship where the subject organization is a member (or part) of the object organization, and its inverse.
% Their sole distinction lies in the fact that \texttt{ORGANIZATION:MEMBER\_OF} indicates an {\em autonomous} relationship between the subject and the object (i.e., the subject is a member of the object by choice), while \texttt{ORGANIZATION:PARENTS} indicates a dependent link where the subject is subsumed by the object (e.g., ``\texttt{[LinkedIn]\textsubscript{SUB}}'' and ``\texttt{[Microsoft]\textsubscript{OBJ}}''), and similarly for the second pair.
% While such fine-grained distinctions may be viable in a document-level relation extraction setting---The TAC KBP evaluations were defined as document-level relation extraction tasks---they can be extremely challenging (even impossible) at the sentence-level, where significantly less information is available.
% In fact, in multiple of the cases that we manually reviewed, the correct label could only be determined through a search on the Internet, rather than by relying on the provided sentences.
% Thus, we decided to merge the two pairs of relations into \texttt{ORGANIZATION:MEMBER\_OF} and \texttt{ORGANIZATION:MEMBERS}, respectively.
The relations \texttt{ORGANIZATION:MEMBER\_OF} and \texttt{ORGANIZATION:PARENTS} describe the relationship where a subject organization is a member (or part) of an object organization.
Their sole distinction lies in the fact that \texttt{ORGANIZATION:MEMBER\_OF} indicates an {\em autonomous} relationship between the subject and the object (i.e., the subject is a member of the object by choice), while \texttt{ORGANIZATION:PARENTS} indicates a dependent link where the subject is subsumed by the object (e.g., ``\texttt{[LinkedIn]\textsubscript{SUB}}'' and ``\texttt{[Microsoft]\textsubscript{OBJ}}'').
While such fine-grained distinctions may be viable in a document-level relation extraction setting, such as that of the TAC KBP evaluations, they can be extremely challenging (if not impossible) at the sentence-level, where significantly less information is available.
In fact, in multiple cases that we manually reviewed, the correct label could only be determined through a search on the Internet, rather than by relying on the provided sentences.
Thus, we merged these two relations into one: \texttt{ORGANIZATION:MEMBER\_OF}. 
Additionally, we similarly merged their inverses, \texttt{ORGANIZATION:MEMBERS} and \texttt{ORGANIZATION:SUBSIDIARIES}, into \texttt{ORGANIZATION:}- \texttt{MEMBERS}.

\vspace{1ex}\noindent\underline{Single-Label vs Multi-Label:}
Although TACRED is defined as a single-label relation extraction dataset (i.e., the relations are all {\em mutually-exclusive}), certain sentences can fit multiple relations.
% the TAC KBP evaluation allows multiple labels between any subjects and objects. 
% is defined as a multi-label relation extraction task.
% As a result, there exist a few cases where multiple relations may apply to a single sentence.
This is especially common among sentences which invoke a residential relationship between people and locations.
For example, both relations \texttt{PERSON:CITIES\_OF\_RESIDENCE} and \texttt{PERSON:CITY\_OF\_BIRTH} apply to the sentence ``\texttt{[He]\textsubscript{SUB} is a native of [Potomac]\textsubscript{OBJ}, Maryland}.''
We account for these cases by altering the relation definitions to create clear boundaries for when one relation is more appropriate over another (e.g., any mention of the word ``native'' or any of its synonyms cannot be assigned a residence relation, such as \texttt{PERSON:CITIES\_OF\_RESIDENCE}).

\vspace{1ex}\noindent\underline{\texttt{ORGANIZATION:*\_OF\_HEADQUARTERS}:}
We also made alterations to the \texttt{ORGANIZATION:*\_OF\_HEADQUARTERS} relations, where ``\texttt{*}'' is a placeholder for location types (e.g., \texttt{CITY}).
Our initial annotation process for these relations resulted in substantial confusion due to semantic ambiguities present throughout the data.
For example, does the phrase ``\texttt{ORGANIZATION from CITY}'' always imply that the specified organization is headquartered in the specified city?
Based on the TAC KBP guidelines it may, but determining whether it does turned out to be particularly challenging for our annotators.
Based on this observation, we generalized the corresponding relation definitions to be valid as long as an organization has a branch or office in the label's location type (rather than forcing it to be headquartered there). 
For example, \texttt{ORGANIZATION:CITY\_OF\_HEADQUARTERS} became \texttt{ORGANIZATION:CITY\_OF\_BRANCH}. 

\subsection{Quality Assurance}
\label{sec:quality-assurance}

In order to ensure high-quality annotations, we employed a two-step quality assurance process similar to the gated-instruction technique introduced by \citet{gated-instructions} for our crowd annotators.
The first step, which we call the {\em trial}, is conducted prior to the data annotation process, and is used to filter out annotators that perform poorly before they are able to label our data.
% starting to annotate our data.
The second stage, which we call the {\em control}, is performed during our data annotation process in order to ensure consistent high-quality annotations.

\vspace{0.65ex}\noindent\underline{Trial:}
We specify several prerequisite criteria that workers must satisfy before annotating our dataset.
First, candidates must have had at least $500$ previous tasks approved on Amazon Mechanical Turk (AMT), and an overall approval rate $\big(\frac{\text{\# Annotations Approved}}{\text{\# Annotations Completed}}\big)$ $\geq 95\%$.
These filters help ensure that our annotators are both experienced and reliable.
In addition, we constructed custom ``qualification tests'' for all eight of our sentence super-clusters.
Since all sentences within a super-cluster are assigned the same set of candidate relations, we made sure that each test contained the definitions of all candidate relations assigned to the respective super-cluster, along with a series of questions aimed at testing a worker's understanding of each of these relations.
A perfect score of 100\% was required to pass.
These tests serve two purposes:
(i) gauge annotator quality, and
(ii) {\em specialize/train} annotators for each super-cluster annotation task.
Only annotators that passed these tests were allowed to provide annotations.

\vspace{0.65ex}\noindent\underline{Control:}
Although our prerequisites were sufficient to eliminate many untrustworthy workers, we observed several incidents where annotators would devote effort to pass our trial criteria, and then randomly annotate sentences to save time while getting paid.
While such events may be easy to detect at small scales where a comprehensive manual review of each annotation is viable, it is infeasible to do so at a large scale involving tens of thousands of sentences.
Thus, we handpicked and manually labeled a set of {\em control} sentences, and mixed them with the unannotated sentences presented to annotators.
% we decided to handpick a set of {\em control} sentences, which we manually annotated ourselves, and mix them with the un-annotated sentences presented to the annotators.
Following the work of \citet{the-crowd}, for every five sentences presented to annotators, we made sure that one was a control sentence whose true label was known.
This allowed us to estimate the annotator accuracy, which in turn enabled us to impose a filter that only accepted responses from annotators with accuracy higher than 80\% (separately computed for each one of our super-clusters). 
We choose this threshold based off that used by \citet{the-crowd} throughout their experiments. 
On average, this eliminated approximately 10\% of the annotators, and significantly improved the quality of the collected data.
Note that, in aggregate we used approximately 2,000 unique control sentences for the annotation of the full TACRED dataset.
% Our control sentences are gathered from a manual annotation over five thousand TACRED sentences. Of these, we kept only those whose labels were unequivocally correct. In aggregate we incorporated two thousand control sentences in our task annotations. Lastly, we reject {\em all} annotations by workers who scored below $80\%$ overall control questions they labeled {\em per} cluster. On average, this eliminated the bottom $10\%$ of annotators, and enabled us to seamlessly remove undesirable workers. Figure 1 illustrates worker performances over our control across our sentence clusters.

\subsection{Miscellaneous Revisions}
\label{sec:miscellaneous-revisions}
% In addition to the aforementioned modifications to the crowdsourcing task design, 
We noticed that 
% In addition to the aforementioned modifications to the crowdsourcing task design, we also discovered a couple of issues with the TACRED datasets that we addressed separately.
% First, we noticed that 
1,058 of the TACRED sentences were not written in English (we automated this detection process by using FastText by \citet{fasttext}).
Since the task is defined in the English language, we removed these sentences from the dataset, leaving us with 105,206 sentences.
% Second, we observed that in over 55.5\% of the sentences, the provided named-entity recognizer (NER) tags conflicted with the specified subject and object types, which were supposedly derived from the NER tags.
% In fact, in some sentences, the subject and object types were even reversed in the NER tags.
% We addressed this problem by replacing the subject and object NER tags with their respective types.

% \begin{figure*}[t]
%     \centering
%     \includegraphics[width=.95\textwidth]{LaTeX/figures/ner_conflict_fig.png}
%     \caption{Example where NER tags conflict.}
%     \label{fig:jrrelp-overview}
%     \vspace{-4ex}
% \end{figure*}

% \subsection{Sentence Pruning}
% \label{sec:sentence_pruning}
Additionally, we analyzed the sentences which gave our workers the most difficulty after finishing our crowd annotation.
We defined difficulty according to the proportion of disagreement between workers. 
Closely inspecting a random sample of 500 sentences, we found that difficulties predominately arose from entities whose spans only partially describe objects. 
For instance, consider the phrase, ``\texttt{[Champions]\textsubscript{OBJ}} League''.
Despite the phrase referring to a European sports league, the given object span creates substantial ambiguity---does \texttt{Champions} refer to the league, or a group of people? 
Due to such ambiguities, we opted to remove such sentences from the dataset, leaving us with 91,467 sentences.

\section{TACRED and Re-TACRED Comparison}
\label{sec:analysis}

After revision, Re-TACRED consists of 91,467 sentences split amongst 40 different relations. 
In order to maintain a similar evaluation environment as TACRED, Re-TACRED contains the same train, development, and test splits as TACRED.
In this section we first provide a qualitative comparison between TACRED and 
% our re-annotated version, 
Re-TACRED.
Then, we provide an empirical analysis for how our re-annotation efforts affect model performance and potentially influence conclusions that were previously drawn from their TACRED evaluations.

\subsection{Qualitative Comparison} \label{sec:qual-comp}

Overall, our Re-TACRED labels achieved an average agreement rate of $82.3\%$ between annotators throughout the whole dataset. %, suggesting both a high degree of task and label clarity.
Moreover, our inter-annotator Fleiss' Kappa over all annotations is .77, indicating high quality.
Our labels disagree with the original TACRED labels in 23.9\% of sentences.
% Perhaps most importantly, when \citet{palstm} originally published TACRED they chose a random sample of 300 sentences and estimated that the error rate in their annotations was around 6.7\%
% but our re-annotation effort and subsequent analysis indicate that the TACRED annotations error rate is in fact approximately 23.9\%.
% This number is computed based on how many of the TACRED labels disagree with the Re-TACRED labels, which we consider more accurate overall because of both the improved task design and the significantly higher inter-annotator agreement rate.
Out of the modified labels, 75.3\% correspond to \texttt{NO\_RELATION} that are switched to one of the other relations and 16.1\% correspond to other relations switching to \texttt{NO\_RELATION}.
The remaining 8.6\% correspond to switching between different non-negative relations.
Our revisions also substantially alter the distribution of relations in TACRED.
For instance, we observed that 41.8\% more sentences are labeled with \texttt{PERSON:CITY\_OF\_BIRTH} than in the original dataset.
Of these, 55.2\% were originally labeled as \texttt{PERSON:CITIES\_OF\_RESIDENCE}, illustrating the effect of improved label definitions at defining concrete bounds between the two relations.
Moreover, we observed a 67.5\% average increase in labels describing organizations in locations.
% (e.g., \texttt{ORGANIZATION:CITY\_OF\_HEADQUARTERS}).
Of these revisions, 93.9\% were originally labeled as \texttt{NO\_RELATION}.
We attribute this influx of assignments primarily due to our changes in the respective relation definitions described in Section~\ref{sec:relation-definitions-refinement}, as well as our efforts to better handle wrong assignments of subject and object types.

While our revisions increase the presence of many labels, they also substantially decrease the presence of several others.
For instance, we observed the largest reduction in \texttt{PERSON:CITIES\_OF\_RESIDENCE}, where 44.6\% of the sentences were re-annotated with a different label.
Interestingly, this complements our aforementioned increase in sentences labeled with \texttt{PERSON:CITY\_OF\_BIRTH}, suggesting a high rate of confusion between the two in the original TACRED dataset.
This pattern is also mirrored for the \texttt{PERSON:COUNTRIES\_OF\_RESIDENCE} and \texttt{PERSON:STATES\_OR\_PROVINCES\_OF\_RESIDENCE} relations which changed to the \texttt{PERSON:COUNTRIES\_OF\_BIRTH} relation and the \texttt{PERSON:STATES\_OR\_PROVINCES\_OF\_BIRTH} relation, respectively.
Additionally, we found a 40.1\% decrease in sentences labeled with the \texttt{PERSON:OTHER\_FAMILY} relation.
We attribute this decrease due to our addition of the \texttt{PERSON:IDENTITY} relation.

\subsection{Model Performance Comparison}
\label{sec:model-performance}

% In addition to analyzing what changed between TACRED and Re-TACRED in terms of the sentence labels we also perform an analysis
% over how these changes affect model performance.
We examine how our changes impact the evaluation of three existing relation extraction models (neither of which are our own):
% and discuss the conclusions reached based on that evaluation.
% In addition to analyzing label distributional changes between TACRED and our revised dataset, we also examine how our labels impact existing model performances. Specifically, we explore two broad questions. First, how much does {\em annotation revision} contribute to performance difference? Second, how much do methods themselves factor into these changes?
% Specifically, we perform an analysis using three existing relation extraction models (neither of which are our own):
\begin{itemize}[noitemsep,label=--]
    \item \underline{PA-LSTM (\citet{palstm}):}
        This model infers relations by applying a one-directional long short-term memory (LSTM) network and a custom position-aware attention mechanism over sentences.
        It also incorporates sentence token named-entity recognition (NER) tags, part-of-speech (POS) tags, and positional offsets from subjects and objects in its reasoning.
        We refer readers to \citet{palstm} for further information.
    \item \underline{C-GCN (\citet{cgcn}):}
        This model labels sentences by applying a graph-convolution network (GCN) over sentence dependency tree parses.
        Similar to PA-LSTM, the model first encodes sentences using a bi-directional LSTM network, before processing the outputs over a graph implied by a pruned version of the sentence dependency tree parse.
        In particular, C-GCN computes the least common ancestor (LCA) between the subject and the object, and removes tree branches that are more than a pre-specified degree away from the LCA.
        The resulting GCN output representations are finally processed by a multi-layer perceptron to predict relations.
        We refer readers to \citet{cgcn} for further information.
    \item \underline{SpanBERT (\citet{spanbert}):}
        This is one of the state-of-the-art models at the time of writing. SpanBERT is similar to BERT (\citet{bert}), but is instead pre-trained using a span prediction objective, making it better suited to the relation extraction task.
        SpanBERT also differs from BERT in terms of how the token masking is performed during pre-training, in that it masks contiguous token spans instead of individual tokens.
        We refer readers to \citet{spanbert} for further information.
        % Note that, to our knowledge, this is the best performing relation extraction model that is also open sourced (and we can thus use in our evaluation).
        % We choose this method because it is the best performing model that does not leverage external data \cite{bert-em,knowbert} and for which code is publicly available.
\end{itemize}

\begin{table}[!t]
    \centering
    \scriptsize
    \small
	\setlength{\tabcolsep}{2pt}
    \renewcommand{\arraystretch}{1.}
	\begin{tabularx}{\columnwidth}{|l|X|Y|Y|Y|}
		\hhline{|-|-|-|-|-|}
		\multirow{2}{*}{\textbf{Dataset}} & \multirow{2}{*}{\textbf{Metric}} & \multicolumn{3}{c|}{\textbf{Models}} \\
% 		\vskip .1cm
		\hhline{|~|~|---|}
		& & \textbf{\mbox{PA-LSTM}} & \textbf{C-GCN} & \textbf{SpanBERT} \\
		\hhline{:=:=:===:}
		\multirow{3}{*}{TACRED}    & Precision & 68.1 & 68.5 & 70.1 \\
		                           & Recall    & 64.5 & 64.4 & 69.2 \\
		                           & F1        & 66.2 & 66.3 & 69.7 \\
		\hhline{|-|-|---|}
		\multirow{3}{*}{Re-TACRED} & Precision & 79.2 & 80.9 & \textbf{85.2} \\
		                           & Recall    & 79.5 & 79.7 & \textbf{85.4} \\
		                           & F1        & 79.4 & 80.3 & \textbf{85.3} \\
		\hhline{|-|-|---|}
		\multirow{3}{*}{Difference} & Precision & +11.1 & +12.4 & \textbf{+15.1} \\
		                           & Recall    & +15.0 & +15.3 & \textbf{+16.2} \\
		                           & F1        & +13.2 & +14.0 & \textbf{+15.6} \\
		\hhline{|-|-|---|}
	\end{tabularx}
    \caption{
        Results for multiple RE models.
        We report result for TACRED obtained using our own experiments that may differ slightly from previously reported numbers.
        ``Difference'' indicates the performance difference between methods evaluated on TACRED and Re-TACRED.
    }
    \label{tab:results}
\end{table}

\vspace{1ex}\noindent\textbf{Overall Performance Impact.}
Table~\ref{tab:results} presents the evaluation results of the three models on both TACRED and Re-TACRED. 
In addition, we mark their performance differences between the two datasets.
% We present the evaluation results of the three models when using both TACRED and Re-TACRED datasets in Table~\ref{tab:results}.
% In addition, we record their performance differences between the two datasets.
% In addition, we record the improvement percentages of models evaluated on Re-TACRED have over those assessed on TACRED. 
All results were reported using micro-averaged f1-scores from the model with the median validation f1-score over five independent runs, as in prior literature.  
Interestingly, while C-GCN is marginally better than PA-LSTM in TACRED, their differences are more pronounced in Re-TACRED: C-GCN outperforms PA-LSTM by as much as $1.7\%$ in Precision. 
% Interestingly, although PA-LSTM and C-GCN have similar f1-score increases, their recall and precision enhancements are complementary. C-GCN has larger recall improvement, while PA-LSTM displays a larger precision increase. 
Notably, we observe significant improvements across every metric for each of the three models. SpanBERT achieves the largest improvement in f1-measure by $15.6\%$,  precision by $15.1\%$, and a $16.2\%$ improvement in recall.
These asymmetric model behavior differences indicate that improvement is not simply due to a revision offset or score scaling; instead, it is dependent on the characteristics of each model at reasoning over diverse data. In addition, these results suggest that existing models are under-evaluated on TACRED, and that their true capabilities---and performance margins---may be significantly better than reported.

\begin{table*}[!t]
    \centering
    \scriptsize
    \small
	\setlength{\tabcolsep}{3pt}
    \renewcommand{\arraystretch}{.9}
	\begin{tabularx}{\textwidth}{|l|l|YYYYYYY|}
		\hhline{|-|-|-------|}
		\multirow{2}{*}{\textbf{Model}} & \multirow{2}{*}{\textbf{Dataset}} & \multicolumn{7}{c|}{\textbf{Categories}} \\
		\hhline{|~|~|-------|}
		& & \texttt{PER:*} & \texttt{ORG:*} & \texttt{PER:ORG} & \texttt{ORG:PER} & \texttt{PER:LOCATION} & \texttt{PER:PER} & \texttt{ORG:ORG} \\
		\hhline{:=:=:=======:}
		\multirow{3}{*}{PA-LSTM} 
        & \texttt{TACRED}    & \phantom{0}66.8 & \phantom{0}65.2 & \phantom{0}65.3 & \phantom{0}72.6 & \phantom{0}51.9 & \phantom{0}59.9 & \phantom{0}59.3 \\
        & \texttt{Re-TACRED} & \phantom{0}79.0 & \phantom{0}74.4 & \phantom{0}68.3 & \phantom{0}85.1 & \phantom{0}53.4 & \phantom{0}85.2 & \phantom{0}70.3 \\
        %  & \texttt{Re-TACRED} & \phantom{0}80.4 & \phantom{0}75.7 & \phantom{0}68.3 & \phantom{0}82.8 & \phantom{0}51.5 & \phantom{0}86.0 & \phantom{0}72.7 \\
        & \texttt{Difference} & +12.2 & \phantom{0}+9.2 & \phantom{0}+3.0 & +12.5 & \phantom{0}+1.5 & +15.3 & +11.0 \\
        % & \texttt{Difference} & +13.6 & \phantom{0}+10.5 & \phantom{0}+3.0 & +10.2 & \phantom{0}-0.4 & +16.1 & +13.4 \\
        \hhline{|-|-|-------|}
        \multirow{3}{*}{C-GCN} 
        & \texttt{TACRED}    & \phantom{0}66.5 & \phantom{0}65.9 & \phantom{0}66.4 & \phantom{0}72.2 & \phantom{0}51.5 & \phantom{0}49.9 & \phantom{0}61.6 \\
        & \texttt{Re-TACRED} & \phantom{0}81.0 & \phantom{0}78.1 & \phantom{0}69.0 & \phantom{0}86.8 & \phantom{0}55.1 & \phantom{0}86.2 & \phantom{0}73.8 \\
        & \texttt{Difference} & +14.5 & +12.2 & \phantom{0}+2.6 & +14.6 & \phantom{0}+3.6 & +36.3 & +12.2 \\
        \hhline{|-|-|-------|}
        \multirow{3}{*}{SpanBERT} 
        & \texttt{TACRED}    & \phantom{0}69.7 & \phantom{0}69.5 & \phantom{0}68.9 & \phantom{0}74.8 & \phantom{0}55.9 & \phantom{0}61.2 & \phantom{0}68.1 \\
        & \texttt{Re-TACRED} & \phantom{0}85.9 & \phantom{0}83.3 & \phantom{0}80.6 & \phantom{0}88.8 & \phantom{0}71.2 & \phantom{0}88.6 & \phantom{0}82.4 \\
        & \texttt{Difference} & +16.2 & +13.8 & +11.7 & +14.0 & +15.3 & +17.4 & +14.3 \\
        \hhline{|-|-|-------|}
	\end{tabularx}
    \caption{
        Micro-averaged f1-score for each category in {\small TACRED} and {\small Re-TACRED}, along with their percent differences.
        \texttt{PER} stands for \texttt{PERSON} and \texttt{ORG} for \texttt{ORGANIZATION}. ``\texttt{*}'' indicates all object types.
    }
    \label{tab:category_results}
\end{table*}

\begin{table*}[!t]
    \centering
    \scriptsize
    \small
	\setlength{\tabcolsep}{1pt}
    \renewcommand{\arraystretch}{1.}
	\begin{tabularx}{\textwidth}{|l|l|YYYYYYY|}
		\hhline{|-|-|-------|}
		\multirow{2}{*}{\textbf{Model}} & \multirow{2}{*}{\textbf{Dataset}} & \multicolumn{7}{c|}{\textbf{Refined Labels}} \\
		\hhline{|~|~|-------|}
		& & \texttt{\textbf{ORG:MEMBER\_OF}} & \texttt{\textbf{ORG:MEMBERS}} & \texttt{\textbf{PER:RESIDENCE}} & \texttt{\textbf{PER:BIRTH}} & \texttt{\textbf{PER:DEATH}} & \texttt{\textbf{ORG:LOCATION}} & \texttt{PER:IDENTITY} \\
		\hhline{:=:=:=======:}
        \multirow{3}{*}{PA-LSTM} 
        & \texttt{TACRED}     & \phantom{+}22.6 & \phantom{+}23.5 & \phantom{+}54.1 & \phantom{+}31.0 &  \phantom{+}26.7 &  \phantom{+}55.9 &  \phantom{+0}0.0 \\
        & \texttt{Re-TACRED}  & \phantom{+}42.7 &   \phantom{+}48.8 &  \phantom{+}55.2 & \phantom{+}57.1 & \phantom{+}40.5 &  \phantom{+}68.1 &  \phantom{+}87.8 \\
        % & \texttt{Re-TACRED}  &  \phantom{+}54.9 &  \phantom{+}41.7 & \phantom{+}52.9 & \phantom{+}64.9 &  \phantom{+}37.9 &  \phantom{+}71.5 &  \phantom{+}88.8 \\
        & \texttt{Difference} & +20.1 & +25.3 & \phantom{0}+1.1 & +26.1 & +13.8 & +12.2 & +87.8 \\
        \hhline{|-|-|-------|}
        \multirow{3}{*}{C-GCN} 
        & \texttt{TACRED}     &  \phantom{+}24.6 &  \phantom{+}24.0 &  \phantom{+}54.1 &  \phantom{+}30.0 &  \phantom{+}25.0 &  \phantom{+}56.7 & \phantom{+}\phantom{0}0.0 \\
        & \texttt{Re-TACRED}  &  \phantom{+}43.1 &  \phantom{+}61.8 &  \phantom{+}55.8 &  \phantom{+}56.4 &  \phantom{+}49.4 &  \phantom{+}74.1 &  \phantom{+}88.0 \\
        & \texttt{Difference} & +18.5 & +37.8 & \phantom{0}+1.7  & +26.4 & +24.4 & +17.4 & +88.0 \\
        \hhline{|-|-|-------|}
        \multirow{3}{*}{SpanBERT} 
        & \texttt{TACRED}     &  \phantom{+}50.3 &  \phantom{+}52.2 &  \phantom{+}57.6 &  \phantom{+}50.0 &  \phantom{+}26.7 &  \phantom{+}62.7 &  \phantom{+}19.1 \\
        & \texttt{Re-TACRED}  &  \phantom{+}73.4 &  \phantom{+}70.0 &  \phantom{+}70.0 &  \phantom{+}81.8 &  \phantom{+}74.5 &  \phantom{+}78.5 &  \phantom{+}90.5 \\
        & \texttt{Difference} & +23.1 & +17.8 & +12.4 & +31.8 & +47.8 & +15.8 & +71.4 \\
        \hhline{|-|-|-------|}
	\end{tabularx}
    \caption{
        Micro-averaged f1-score for all our refined labels in {\small TACRED} and {\small Re-TACRED}, along with their percent differences.
        \texttt{PER} stands for \texttt{PERSON}, and \texttt{ORG} stands for \texttt{ORGANIZATION}. 
        The refined relations are grouped according to their type, and are defined as in Section~\ref{sec:relation-definitions-refinement}.
        Additionally, \texttt{PER:RESIDENCE}, \texttt{PER:BIRTH}, and \texttt{PER:DEATH} represent all \texttt{LOCATION} types of residence, birth, and death respectively, for type \texttt{PER}.
        \texttt{ORG:LOCATION} is the aggregate of all \texttt{LOCATION} types for \texttt{ORG} subjects.
        \texttt{PERSON:IDENTITY} refers to \texttt{PERSON:ALTERNATE\_NAMES} in TACRED evaluations.
    }
    \label{tab:def_impact}
\end{table*}

\vspace{1ex}\noindent\textbf{Performance Change Across Label Types.}
To better understand these performances, we also analyze model quality over several relation categories. Each category examines particular relation types, and is defined similar to \citet{tacredReannotate}. Namely, \texttt{PER:*} and \texttt{ORG:*} represent all relations whose subject types are \texttt{PERSON} and \texttt{ORGANIZATION} respectively, while those denoted by \texttt{X:Y} symbolize relations whose subject type is \texttt{X} and object type is \texttt{Y}. We choose these categories due to the diversity of specific relations they represent, and their overall coverage of the relation-space. For each category, we compute the micro-averaged f1-score based on the scores from its relations. We report our results in Table \ref{tab:category_results}. 

The results indicate that C-GCN and PA-LSTM exhibit a complementary relationship over many categories with TACRED labels. While C-GCN beats PA-LSTM in \texttt{ORGANIZATION:*}, the reverse is true with \texttt{PERSON:*}. Moreover, PA-LSTM significantly outperforms C-GCN by $10\%$ on \texttt{PERSON:PERSON} relationships. However, this relationship disappears when the two are compared on our revised dataset. Notably, C-GCN outscores PA-LSTM in every category. Thus, while TACRED paints these methods as being comparable, Re-TACRED reveals that C-GCN is a much stronger model.  
SpanBERT consistently beats PA-LSTM and C-GCN in both TACRED and Re-TACRED evaluations, illustrating its robustness.

% While these results echo our previous observations that Re-TACRED significantly improves existing model results, they also present a completely different view of model differences. For instance, C-GCN and PA-LSTM exhibit a complementary relationship over many categories with TACRED labels. While C-GCN edges PA-LSTM in \texttt{org:*} relations, the reverse is true with \texttt{per:*} labels. Moreover, PA-LSTM significantly outperforms C-GCN by by $12.5\%$ in \texttt{per:per} relationships.
% However, this relationship disappears when the two are compared on the revised dataset. Notably, C-GCN outscores PA-LSTM in almost category, at worst equaling PA-LSTM in the \texttt{per:per} group.
% SpanBERT also reveals similar patterns. While performing better than C-GCN and PA-LSTM in many TACRED categories, the model scores substantially lower in both \texttt{per:org} and \texttt{org:per} groups. However, SpanBERT {\em significantly outperforms every method in every category} on the revised dataset.

\vspace{1ex}\noindent\textbf{Effect of Refined Labels.} We also examine how impactful our label refinements are across different models. Table \ref{tab:def_impact} reports the micro-averaged f1-scores for each label refinement-category on TACRED and Re-TACRED. Categories are defined as in Section \ref{sec:relation-definitions-refinement}, with a few additions. 
Namely, we group all \texttt{PERSON}, \texttt{RESIDENCE}, \texttt{BIRTH}, and \texttt{DEATH} types into respective \texttt{PERSON:RESIDENCE}, \texttt{PERSON:BIRTH}, and \texttt{PERSON:DEATH} categories. 
In a similar manner, \texttt{ORGANIZATION:LOCATION} marks all location-type relations (e.g., \texttt{ORGANIZATION:CITY\_OF\_BRANCH}) describing the place of an \texttt{ORGANIZATION}'s branch or office. 
% \texttt{REST} denotes the set of all remaining labels. 

Overall, our label refinements yield significant performance improvements across all models {\em by as much as  88.0\%}. While PA-LSTM and C-GCN performances are difficult to distinguish on TACRED, C-GCN exhibits better performance than PA-LSTM after label refinement.
Similarly, SpanBERT achieves significantly better f1-scores, by an average of $31.4\%$ across categories.
% at least $12.4\%$ in every category. 
Its best improvement is on \texttt{PERSON:IDENTITY}, showing a $71.4\%$ increase in f1-measure, highlighting the added clarity of our label refinements.
Moreover, all methods achieve the largest gain in \texttt{PERSON:IDENTITY} classifications, and two---PA-LSTM and C-GCN---improve performance from 0.0\% to more than 87.0\%. This indicates that their robustness is at detecting same-person relationships is significantly higher than could be observed in TACRED.
Interestingly, all models exhibit the least improvement on \texttt{PERSON:RESIDENCE} labels. We hypothesize that this is because their relations are more much more complex than similar labels such as birth and death. Specifically, whereas lexical variation describing places of birth and death is limited, characterizations of locations of residences are diverse in the TAC KBP documentation. For instance, ``grew up", ``lives", ``has home", ``from", etc$\hdots$ are just a few of many valid indications. Moreover, we observe substantial improvements in \texttt{ORGANIZATION:MEMBERS} and \texttt{ORGANIZATION:MEMBER\_OF}. Both categories yielded among the lowest scores for models evaluated on TACRED, illustrating their difficulties in distinguishing between the subtle label differences in each group. By addressing these nuances, we observe significant f1-score increase on Re-TACRED.

\begin{table*}[!t]
    \centering
    \scriptsize
    \setlength{\tabcolsep}{3pt}
    \renewcommand{\arraystretch}{0.8}
    \begin{tabularx}{\textwidth}{|c|l|ll|}
    \hhline{|-|-|--|}
    \textbf{Error Type} & \textbf{Sentence} & \textbf{TACRED Prediction} & \textbf{Correct Label} \\
    \hhline{:=:=:==:}
    \multirow{3}{*}{\texttt{Neg $\rightarrow$ Pos}} 
    & ``\ldots \texttt{[Motorola]\textsubscript{OBJ}} (where \texttt{[he]\textsubscript{SUB}} was also a VP) \ldots" & \texttt{NO\_RELATION} & \texttt{PERSON:EMPLOYEE\_OF} \\
    % ``\ldots Ouattara's newly appointed \texttt{[UN]\textsubscript{OBJ}} envoy, \texttt{[Youssoufou Bamba]\textsubscript{SUB}} \ldots" & \texttt{NO\_RELATION} & \texttt{PERSON:EMPLOYEE\_OF} \\
    % \hhline{|-|--|}
    & ``\texttt{[Filmaker]\textsubscript{OBJ}} and attorney \texttt{[Sarah Kunstler]\textsubscript{SUB}} \ldots" & \texttt{NO\_RELATION} & \texttt{PERSON:TITLE} \\
    % \hhline{|-|--|}
    & ``\ldots \texttt{[National Taiwan Symphony Orchestra]\textsubscript{SUB}} (NTSO) \ldots an \texttt{[NTSO]\textsubscript{OBJ}} spokesman\ldots" & \texttt{NO\_RELATION} & \texttt{ORGANIZATION:ALTERNATE\_NAMES} \\
    \hhline{|-|-|--|}
    \texttt{Pos $\rightarrow$ Neg} & ``\texttt{[His]\textsubscript{SUB}} \texttt{[therapist]\textsubscript{OBJ}} told him to politely decline, ‘which helped." & \texttt{PERSON:TITLE} & \texttt{NO\_RELATION} \\
    \hhline{|-|-|--|}
    \texttt{Pos $\rightarrow$ Pos} & ``\ldots \texttt{[her]\textsubscript{SUB}} stepchildren, Susan, \ldots, Stephen and \texttt{[Maggie]\textsubscript{OBJ}} Mailer; \ldots.'' & \texttt{PERSON:SIBLINGS} & \texttt{PERSON:CHILDREN} \\

    \hhline{|-|-|--|}
    \end{tabularx}
    \caption{
    Five handpicked sentences from the Re-TACRED test split that a TACRED-trained SpanBERT model misclassifies but a Re-TACRED-trained SpanBERT method correctly classifies. Sentence subjects and objects are defined as in Section \ref{sec:introduction}, and the complete TACRED-trained SpanBERT predictions and gold labels are provided. Additionally, each sentence is marked by a specific ``error type'' (the leftmost column) decribing whether the error is due to predicting a negative sentence label when the correct relation is positive (\texttt{Neg $\rightarrow$ Pos}), predicting a positive relation when the correct label is negative (\texttt{Pos $\rightarrow$ Neg}), or inferring the incorrect positive label (\texttt{Pos $\rightarrow$ Pos}).
    }
    \label{tab:tacred_sentence_corrections}
\end{table*}

\vspace{1ex}\noindent\textbf{Effect of Non-Refined Labels.} We also examine how models differ based on our {\em non-refined} label re-annotations. 
Non-refined relations are any for which we did not alter the TAC KBP relation definitions for (i.e. any label not discussed in Section \ref{sec:relation-definitions-refinement}).
We conduct this analysis by comparing model performance over different combinations of train and test splits from TACRED and Re-TACRED.
We denote train splits using {[$\cdot$]\textsubscript{\texttt{train}}} and test splits using {[$\cdot$]\textsubscript{\texttt{test}}}, where [$\cdot$] is either {TACRED} or {Re-TACRED} (e.g., TACRED\textsubscript{\texttt{train}}).
All models are then trained on TACRED\textsubscript{\texttt{train}} or Re-TACRED\textsubscript{\texttt{train}}, and evaluated on TACRED\textsubscript{\texttt{test}} or Re-TACRED\textsubscript{\texttt{test}}.
% Specifically, methods are trained on the training split of one dataset, and evaluated on the test split of another. For instance, one such combination may be training on TACRED, and then tested on Re-TACRED. 
Our results are shown in Table~\ref{tab:non-refined_results}.

\begin{table}[!t]
    \centering
    \scriptsize
	\setlength{\tabcolsep}{3pt}
    \renewcommand{\arraystretch}{0.8}
	\begin{tabularx}{\columnwidth}{|l|l|Y|Y|Y|}
		\hhline{|-|-|-|-|-|}
		\multirow{2}{*}{\textbf{Model}} & \multirow{2}{*}{\textbf{(Train Split, Test Split)}} & \multicolumn{3}{c|}{\textbf{Metrics}} \\
		\hhline{|~|~|---|}
		& & \textbf{F1} & \textbf{Precision} & \textbf{Recall} \\
		\hhline{:=:=:===:}
		\multirow{4}{*}{\texttt{PA-LSTM}}   & (\tacredtrain, \tacredtest)      & 72.3 & 71.3 & 73.3 \\
        		                           & (\tacredtrain, \retacredtest)    & 74.8 & 80.1 & 70.2 \\
        		                           & (\retacredtrain, \tacredtest)    & 65.9 & 59.4 & 74.2 \\ 
        		                           & (\retacredtrain, \retacredtest)  & 77.6 & 77.1 & 78.1 \\
		\hhline{|-|-|---|}
		\multirow{4}{*}{\texttt{C-GCN}}    & (\tacredtrain, \tacredtest)      & 72.6 & 71.1 & 74.3 \\
        		                           & (\tacredtrain, \retacredtest)    & 74.8 & 79.6 & 70.6 \\
        		                           & (\retacredtrain, \tacredtest)    & 69.5 & 64.1 & 75.8 \\ 
        		                           & (\retacredtrain, \retacredtest)  & 79.6 & 81.2 & 78.2 \\
		\hhline{|-|-|---|}
		\multirow{4}{*}{\texttt{SpanBERT}} & (\tacredtrain, \tacredtest)      & 75.0 & 74.7 & 75.3 \\ %<-- TODO
        		                           & (\tacredtrain, \retacredtest)    & 78.9 & 84.9 & 73.7 \\ %<-- TODO
        		                           & (\retacredtrain, \tacredtest)    & 72.2 & 66.3 & 79.3 \\
        		                           & (\retacredtrain, \retacredtest)  & 84.9 & 85.1 & 84.7 \\
		\hhline{|-|-|---|}
	\end{tabularx}
    \caption{
        Results for multiple RE models (leftmost column) on different train-and-evaluation combinations. Each combination is represented by a pair of the form (``train split'', ``test split''). For instance, (TACRED\textsubscript{train}, Re-TACRED\textsubscript{test}) indicates that a method is trained on the TACRED train partition and evaluated on the Re-TACRED test split. The remaining columns show metric results. 
        % The leftmost column describes each method tests. The second column describes the train-and-evaluation strategy. 
        % Models are trained on the training split of datasets to the left of the arrow, and evaluated on the test splits of datasets on the right (e.g., TACRED, Re-TACRED indicates that a method is trained on the TACRED train partition evaluated on the Re-TACRED test split. The remaining columns show metric results. 
    }
    \label{tab:non-refined_results}
\end{table}

The results show several interesting differences between TACRED and Re-TACRED.
First, all methods trained and evaluated on TACRED obtain significantly higher performance on the non-refined labels than over the full label set. We attribute this increase to the fact that these relations are less ambiguous compared than the refined ones. 
Second, methods trained on TACRED\textsubscript{train} achieve better performance on Re-TACRED\textsubscript{test} than on TACRED\textsubscript{test}.
This is consistent with the findings in \citet{tacredReannotate}, and suggests that: (i) TACRED may be under-estimating model performance, and (ii) large improvements can be obtained simply by evaluating models on higher quality annotations.
Third, methods trained on Re-TACRED\textsubscript{train} and evaluated on TACRED\textsubscript{test} perform worse than those evaluated on Re-TACRED\textsubscript{test}.
A deeper inspection of the data reveals that such models exhibit significantly fewer correct positively labeled predictions in TACRED\textsubscript{test} than in Re-TACRED\textsubscript{test}, resulting in substantially lower scores. For instance, SpanBERT trained on Re-TACRED\textsubscript{train} exhibits 49.4\% fewer correct positively labeled instances in TACRED\textsubscript{test} compared to Re-TACRED\textsubscript{test}.
% This highlights our label change effects described in 
This highlights the effects of our label changes described in 
Section~\ref{sec:qual-comp}: many positively labeled sentences in Re-TACRED are either negatively labeled or assigned another positive relation in TACRED.
Fourth, models trained and evaluated on Re-TACRED perform significantly better than any other combination.
Thus, while methods trained on TACRED\textsubscript{train} achieve performance boosts when testing on Re-TACRED\textsubscript{test} (compared to evaluating on TACRED\textsubscript{test}), training on Re-TACRED\textsubscript{train} is critical to achieving the strongest performance on Re-TACRED\textsubscript{test}.

\vspace{1ex}\noindent\textbf{Re-TACRED Error Correction.}
We further investigate how model errors change between TACRED and Re-TACRED. We conduct this analysis by training two separate SpanBERT instances on TACRED and Re-TACRED respectively, and evaluate both on the Re-TACRED test split. We then identify which sentences TACRED-trained SpanBERT classifies incorrectly, while SpanBERT trained on Re-TACRED answers correctly.
We choose SpanBERT because it is the best performing model on both TACRED and Re-TACRED out of our three. 
Overall, we find $2,788$ total such sentences. 
Of these, $84.6\%$ are due to TACRED-trained SpanBERT inferring \texttt{NO\_RELATION} when the gold label is positive, $10.0\%$ occur when the model predicts a positive relation when the correct label is negative, and the remaining $5.4\%$ of errors arise when the method classifies the incorrect positive label. 
We argue that TACRED-trained SpanBERT's erroneous \texttt{NO\_RELATION} predictions are primarily due to implicit negative bias TACRED-trained methods have as a result of TACRED's severe \texttt{NO\_RELATION} data skew ($79.6\%$ of sentences are negatively labeled). In contrast, Re-TACRED trained SpanBERT is able to better recognize instances where \texttt{NO\_RELATION)} is not appropriate, potentially due to Re-TACRED containing substantially fewer negatively labeled instances ($63.2\%$). 
Table \ref{tab:tacred_sentence_corrections} 
shows several sentences highlighting the types of prediction errors TACRED-trained SpanBERT makes that Re-TACRED trained SpanBERT is able to correct for.

\section{Conclusion}
We conducted a comprehensive review of the TACRED dataset. We addressed the limitations of previous work by re-annotating the complete dataset using crowdsourcing. Our annotation strategy extended previous studies by accounting for data errors, label definition ambiguity, and annotation quality control. Our results show significantly higher inter-annotator agreement rate and Fleiss' Kappa (.77) than original dataset annotations, suggesting clearer task descriptions and high label annotation reliability. Moreover, we performed a thorough analysis of how existing relation extraction methods compare between datasets, and how errors change between them. Perhaps most notably, we observed an average improvement of $14.3\%$ f1-score from three models on our revised dataset.

\section*{Acknowledgements}
This work has been partly funded by the DARPA Data-Driven Discovery of Models (D3M) program.

\bibliography{references}

\begin{thebibliography}{14}
\providecommand{\natexlab}[1]{#1}
\providecommand{\url}[1]{\texttt{#1}}
\providecommand{\urlprefix}{URL }
\expandafter\ifx\csname urlstyle\endcsname\relax
  \providecommand{\doi}[1]{doi:\discretionary{}{}{}#1}\else
  \providecommand{\doi}{doi:\discretionary{}{}{}\begingroup
  \urlstyle{rm}\Url}\fi

\bibitem[{Alt, Gabryszak, and Hennig(2020)}]{tacredReannotate}
Alt, C.; Gabryszak, A.; and Hennig, L. 2020.
\newblock {TACRED} Revisited: A Thorough Evaluation of the {TACRED} Relation
  Extraction Task.
\newblock In \emph{Proceedings of the 58th Annual Meeting of the Association
  for Computational Linguistics}, 1558--1569. Online: Association for
  Computational Linguistics.
\newblock \doi{10.18653/v1/2020.acl-main.142}.
\newblock \urlprefix\url{https://www.aclweb.org/anthology/2020.acl-main.142}.

\bibitem[{Alt, H{\"u}bner, and Hennig(2019)}]{tre}
Alt, C.; H{\"u}bner, M.; and Hennig, L. 2019.
\newblock Improving Relation Extraction by Pre-trained Language
  Representations.
\newblock In \emph{Automated Knowledge Base Construction (AKBC)}.
\newblock \urlprefix\url{https://openreview.net/forum?id=BJgrxbqp67}.

\bibitem[{Baldini~Soares et~al.(2019)Baldini~Soares, FitzGerald, Ling, and
  Kwiatkowski}]{bert-em}
Baldini~Soares, L.; FitzGerald, N.; Ling, J.; and Kwiatkowski, T. 2019.
\newblock Matching the Blanks: Distributional Similarity for Relation Learning.
\newblock In \emph{Proceedings of the 57th Annual Meeting of the Association
  for Computational Linguistics}, 2895--2905. Florence, Italy: Association for
  Computational Linguistics.
\newblock \doi{10.18653/v1/P19-1279}.
\newblock \urlprefix\url{https://www.aclweb.org/anthology/P19-1279}.

\bibitem[{Chen et~al.(2020)Chen, Hoehndorf, Elhoseiny, and Zhang}]{dgspanbert}
Chen, J.; Hoehndorf, R.; Elhoseiny, M.; and Zhang, X. 2020.
\newblock Efficient long-distance relation extraction with DG-SpanBERT.
\newblock \emph{arXiv:2004.03636 [cs.LG]}
  \urlprefix\url{https://arxiv.org/abs/2004.03636}.
\newblock ArXiv:2004.03636.

\bibitem[{Devlin et~al.(2019)Devlin, Chang, Lee, and Toutanova}]{bert}
Devlin, J.; Chang, M.-W.; Lee, K.; and Toutanova, K. 2019.
\newblock {BERT}: Pre-training of Deep Bidirectional Transformers for Language
  Understanding.
\newblock In \emph{Proceedings of the 2019 Conference of the North {A}merican
  Chapter of the Association for Computational Linguistics: Human Language
  Technologies, Volume 1 (Long and Short Papers)}, 4171--4186. Minneapolis,
  Minnesota: Association for Computational Linguistics.
\newblock \doi{10.18653/v1/N19-1423}.
\newblock \urlprefix\url{https://www.aclweb.org/anthology/N19-1423}.

\bibitem[{Guo, Zhang, and Lu(2019)}]{aggcn}
Guo, Z.; Zhang, Y.; and Lu, W. 2019.
\newblock Attention Guided Graph Convolutional Networks for Relation
  Extraction.
\newblock In \emph{Proceedings of the 57th Annual Meeting of the Association
  for Computational Linguistics}, 241--251. Florence, Italy: Association for
  Computational Linguistics.
\newblock \doi{10.18653/v1/P19-1024}.
\newblock \urlprefix\url{https://www.aclweb.org/anthology/P19-1024}.

\bibitem[{Joshi et~al.(2020)Joshi, Chen, Liu, Weld, Zettlemoyer, and
  Levy}]{spanbert}
Joshi, M.; Chen, D.; Liu, Y.; Weld, D.~S.; Zettlemoyer, L.; and Levy, O. 2020.
\newblock {S}pan{BERT}: Improving Pre-training by Representing and Predicting
  Spans.
\newblock \emph{Transactions of the Association for Computational Linguistics}
  8: 64--77.
\newblock \doi{10.1162/tacl_a_00300}.
\newblock \urlprefix\url{https://www.aclweb.org/anthology/2020.tacl-1.5}.

\bibitem[{Joulin et~al.(2017)Joulin, Grave, Bojanowski, and Mikolov}]{fasttext}
Joulin, A.; Grave, E.; Bojanowski, P.; and Mikolov, T. 2017.
\newblock Bag of Tricks for Efficient Text Classification.
\newblock In \emph{Proceedings of the 15th Conference of the European Chapter
  of the Association for Computational Linguistics: Volume 2, Short Papers},
  427--431. Association for Computational Linguistics.

\bibitem[{Liu et~al.(2016)Liu, Soderland, Bragg, Lin, Ling, and
  Weld}]{gated-instructions}
Liu, A.; Soderland, S.; Bragg, J.; Lin, C.~H.; Ling, X.; and Weld, D.~S. 2016.
\newblock Effective Crowd Annotation for Relation Extraction.
\newblock In \emph{Proceedings of the 2016 Conference of the North {A}merican
  Chapter of the Association for Computational Linguistics: Human Language
  Technologies}, 897--906. San Diego, California: Association for Computational
  Linguistics.
\newblock \doi{10.18653/v1/N16-1104}.
\newblock \urlprefix\url{https://www.aclweb.org/anthology/N16-1104}.

\bibitem[{Peters et~al.(2019)Peters, Neumann, Logan, Schwartz, Joshi, Singh,
  and Smith}]{knowbert}
Peters, M.~E.; Neumann, M.; Logan, R.; Schwartz, R.; Joshi, V.; Singh, S.; and
  Smith, N.~A. 2019.
\newblock Knowledge Enhanced Contextual Word Representations.
\newblock In \emph{Proceedings of the 2019 Conference on Empirical Methods in
  Natural Language Processing and the 9th International Joint Conference on
  Natural Language Processing (EMNLP-IJCNLP)}, 43--54. Hong Kong, China:
  Association for Computational Linguistics.
\newblock \doi{10.18653/v1/D19-1005}.
\newblock \urlprefix\url{https://www.aclweb.org/anthology/D19-1005}.

\bibitem[{Zhang et~al.(2012)Zhang, Niu, R{\'e}, and Shavlik}]{the-crowd}
Zhang, C.; Niu, F.; R{\'e}, C.; and Shavlik, J. 2012.
\newblock Big Data versus the Crowd: Looking for Relationships in All the Right
  Places.
\newblock In \emph{Proceedings of the 50th Annual Meeting of the Association
  for Computational Linguistics (Volume 1: Long Papers)}, 825--834. Jeju
  Island, Korea: Association for Computational Linguistics.
\newblock \urlprefix\url{https://www.aclweb.org/anthology/P12-1087}.

\bibitem[{Zhang, Qi, and Manning(2018)}]{cgcn}
Zhang, Y.; Qi, P.; and Manning, C.~D. 2018.
\newblock Graph Convolution over Pruned Dependency Trees Improves Relation
  Extraction.
\newblock In \emph{Proceedings of the 2018 Conference on Empirical Methods in
  Natural Language Processing}, 2205--2215. Brussels, Belgium: Association for
  Computational Linguistics.
\newblock \doi{10.18653/v1/D18-1244}.
\newblock \urlprefix\url{https://www.aclweb.org/anthology/D18-1244}.

\bibitem[{Zhang et~al.(2017)Zhang, Zhong, Chen, Angeli, and Manning}]{palstm}
Zhang, Y.; Zhong, V.; Chen, D.; Angeli, G.; and Manning, C.~D. 2017.
\newblock Position-aware Attention and Supervised Data Improve Slot Filling.
\newblock In \emph{Proceedings of the 2017 Conference on Empirical Methods in
  Natural Language Processing}, 35--45. Copenhagen, Denmark: Association for
  Computational Linguistics.
\newblock \doi{10.18653/v1/D17-1004}.
\newblock \urlprefix\url{https://www.aclweb.org/anthology/D17-1004}.

\bibitem[{Zhang et~al.(2019)Zhang, Han, Liu, Jiang, Sun, and Liu}]{ernie}
Zhang, Z.; Han, X.; Liu, Z.; Jiang, X.; Sun, M.; and Liu, Q. 2019.
\newblock {ERNIE}: Enhanced Language Representation with Informative Entities.
\newblock In \emph{Proceedings of the 57th Annual Meeting of the Association
  for Computational Linguistics}, 1441--1451. Florence, Italy: Association for
  Computational Linguistics.
\newblock \doi{10.18653/v1/P19-1139}.
\newblock \urlprefix\url{https://www.aclweb.org/anthology/P19-1139}.

\end{thebibliography}

\begin{table*}[!b]
    \centering
    \scriptsize
    \setlength{\tabcolsep}{3pt}
    \renewcommand{\arraystretch}{0.8}
    \begin{tabularx}{\textwidth}{|c|l|X|}
    \hhline{|-|-|-|}
    \textbf{Label Origin} & \textbf{Label} & \textbf{Sentence Definition} \\
    \hhline{:=:=:=:}
    \multirow{3}{*}{\texttt{TAC KBP}} 
    & \texttt{PERSON:ALTERNATE\_NAMES} & Names used to refer to the assigned person that are distinct from the "official" name. \\
    & \texttt{PERSON:OTHER\_FAMILY} & Family other than siblings, parents, children, and spouse (or former spouse). \\
    & \texttt{NO\_RELATION} & There is no relation between the subject and object entity. \\
    
    \hhline{|-|-|-|}
    \texttt{New} & \texttt{PERSON:IDENTITY} & Both entities refer to the same person. \\

    \hhline{|-|-|-|}
    \end{tabularx}
    \caption{
    TAC KBP sentence definitions of \texttt{PERSON:OTHER\_FAMILY},\texttt{PERSON:ALTERNATE\_NAMES}, and \texttt{NO\_RELATION}, and the definition of our new \texttt{PERSON:IDENTITY} label.
    }
    \label{tab:label-definitions}
\end{table*}

\newpage
% \blankpage
% \blankpage
\appendix
\appendixpage

\section{Hyperparameters}
We train all our TACRED-based models using the reported hyperparameters by their respective contributors. All hyperparameter details for our Re-TACRED-based methods can be found below. Additionally, all texttt required to reproduce our results and our new dataset can be found in our repository at \url{https://github.com/gstoica27/Re-TACRED}. We train our PA-LSTM and C-GCN models on a single Nvidia Titan X GPU, and utilized a single Nvidia Tesla V100 GPU to train SpanBERT. 

\vspace{1ex}\noindent\textbf{Re-TACRED PA-LSTM.}
We perform an extensive grid-search over LSTM hidden dimension sizes from \{100, 150, 200, 250, 300\}, LSTM depth of \{1, 2, 3\}, word dropout from \{0.0, 0.01, 0.04, 0.1, 0.25, .5\}, and position-encoding dimension size among \{15, 20, 25, 30, 50, 75, 100\}. However, we observe the best performance with the hyperparameters reported by \citet{palstm}. In addition, we employ the equivalent training strategy as is reported in \citet{palstm} (detailed under Appendix B of their publication). 
%We refer readers to \citet{palstm} for additional information.

\vspace{1ex}\noindent\textbf{Re-TACRED C-GCN.}
Similar to our PA-LSTM experiments, we find that keeping the majority of hyperparemeters equivalent to those reported by \citet{cgcn} yield the best results. The sole parameter we alter is increasing the residual neural network hidden dimension from 200 to 300. In addition, we use the same training procedure as \citet{cgcn} (described in Appendix A of their publication).

\vspace{1ex}\noindent\textbf{Re-TACRED SpanBERT.}
For SpanBERT, we perform a grid-search over learning rate sizes in \{1e-6, 2e-6, 2e-5\} and warm-up proportions in \{.1, .2\}. However, we observe the best performance using the reported parameters by \citet{spanbert}. We refer readers to \citet{spanbert} (detailed in Section 4.2 and Appendix B in their publication) for further details on training strategy.

\section{Amazon Mechanical Turk}
\label{amt}
Our annotation process can be broken down by super-cluster, where each cluster represents as distinct annotation task. 
Overall, we had 8 such tasks (number of super-clusters), and each consisted of 2-4 labeling rounds. 
The first round gathered 2 distinct annotations for every sentence in the respective super-cluster. 
Any disagreements were then given to a third annotator in the second round. 
Then, if necessary, the third and fourth label rounds asked an additional worker to annotate remaining disagreements. 
Sentences to be annotated in label rounds were grouped into Human Intelligence Tasks (HIT). 
Each HIT consisted of 5 sentences, of which one was gold (i.e., its correct label was known). 
We priced HITs competitively at \$.15 per HIT.
We utilized annotations from 243 total workers, and the total time taken to sequentially annotate all our label rounds across all annotation tasks was $\approx$784 hours.
While this number is large, it is important to note that many labeling rounds were completed in parallel, significantly decreasing the overall time.

\section{Relation Alterations}
\label{app:relation-alterations}
% As described in Section \ref{sec:relation-definitions-refinement}, we create the new \texttt{PERSON:IDENTITY} relation to accommodate identity inter-personal relationships. 
As presented in Section \ref{sec:relation-definitions-refinement}, we observed substantial label inconsistencies among sentences describing identity pronominal relationships between entities. 
These sentences were frequently either classified as either \texttt{PERSON:OTHER\_FAMILY} or \texttt{NO\_RELATION} due to the ambiguity of the TAC KBP guidelines. 
In this section we further motivate the creation of the \texttt{PERSON:IDENTITY} relation by describing how the original TAC KBP label documentation is insufficient. 
We do this by presenting the TAC KBP definitions of each affected relation, and explaining their shortcomings.

\subsection{Definitions}
All label definitions are given in Table \ref{tab:label-definitions}. 

\subsection{Encoding Identity Shortcomings}
The TAC KBP relations above cover a wide array of inter-personal relationships between two entities. 
However, classifying identity sentences is notably difficult because neither definition explicitly accounts for the relation. 
While \texttt{PERSON:ALTERNATE\_NAMES} describes two entities that refer to the same person, these entities must be {\em different names}. 
Pronominal reference is not covered in the definition. 
In contrast, \texttt{PERSON:OTHER\_FAMILY} captures general familial relationships. 
Does this include pronominal relations? 
Based on the original TACRED label assignments, the answer to this question appeared subjective:  it depended on the crowd-workers' interpretations. 
If an annotator interpreted pronominal relations as familial, \texttt{PERSON:OTHER\_FAMILY} was chosen, the sentence would be classified as \texttt{NO\_RELATION}.

In an effort to address this issue, we first create the new \texttt{PERSON:IDENTITY} relation (described in Table \ref{tab:label-definitions}), which explicitly covers pronominal relationships between sentence entities. 
Second, because \texttt{PERSON:IDENTITY} is a generalization of \texttt{PERSON:ALTERNATE\_NAMES}, we absorb \texttt{PERSON:ALTERNATE\_NAMES} in \texttt{PERSON:IDENTITY}. 
We then bar \texttt{PERSON:OTHER\_FAMILY} from accepting pronominal relationships while leaving the rest unchanged.

\end{document}